\documentclass[sigconf]{acmart}

\usepackage{amsmath,amsfonts}
\usepackage{graphicx}
\usepackage{comment}
\usepackage{booktabs}
\usepackage{multirow}
\usepackage{subcaption}
\usepackage{eqparbox}
\usepackage{comment}
\usepackage{threeparttable}
\usepackage{xcolor}
\usepackage{algorithm}
\usepackage{mathtools}
\usepackage[compatible]{algpseudocode} 

\renewcommand\footnotetextcopyrightpermission[1]{} 

\usepackage[skip=0pt,font=small]{caption}
\title{RiskLoc: Localization of Multi-dimensional Root Causes by Weighted Risk}
\author{Marcus Kalander}
\affiliation{%
  \institution{Noah's Ark Lab, Huawei Technologies}
  \city{}
  \country{}
}
\email{marcus.kalander@huawei.com}

\begin{document}

\begin{abstract}
Failures and anomalies in large-scale software systems are unavoidable incidents. When an issue is detected, operators need to quickly and correctly identify its location to facilitate a swift repair. In this work, we consider the problem of identifying the root cause set that best explains an anomaly in multi-dimensional time series with categorical attributes. The huge search space is the main challenge, even for a small number of attributes and small value sets, the number of theoretical combinations is too large to brute force. Previous approaches have thus focused on reducing the search space, but they all suffer from various issues, requiring extensive manual parameter tuning, being too slow and thus impractical, or being incapable of finding more complex root causes. 

We propose \textit{RiskLoc} to solve the problem of multidimensional root cause localization. RiskLoc applies a 2-way partitioning scheme and assigns element weights that linearly increase with the distance from the partitioning point. A risk score is assigned to each element that integrates two factors, 1) its weighted proportion within the abnormal partition, and 2) the relative change in the deviation score adjusted for the ripple effect property. 
Extensive experiments on multiple datasets verify the effectiveness and efficiency of RiskLoc, and for a comprehensive evaluation, we introduce three synthetically generated datasets that complement existing datasets.
We demonstrate that RiskLoc consistently outperforms state-of-the-art baselines, especially in more challenging root cause scenarios, with gains in F1-score up to 57\% over the second-best approach with comparable running times.

\end{abstract}

\maketitle
\pagestyle{plain} 

\section{Introduction}
In large software systems, various types of failures and anomalies are common and unavoidable occurrences~\cite{lyu1996handbook}. A failure can severely degrade system performance and affect service availability which may lead to economic losses~\cite{bansal2020decaf}. Therefore, it is critical to quickly identify and localize a failure before it can affect the user experience.
To this end, all important metrics, key performance indicators (KPIs), are generally collected periodically and monitored to identify performance or quality issues~\cite{meng2019loganomaly, zhang2016funnel, zhang2015rapid}.
In practice, anomaly detection models are applied to these collected time series, and when an anomaly is identified, an operator is alerted or the system initializes a self-diagnosis and repair procedure.

However, it is not always feasible or practical to run anomaly detection models on all individual time series due to the overwhelming quantity of KPIs and difficulties with low sample sizes, e.g., sporadic and short-lived user-specific KPIs. 
The anomaly detection is thus often run on aggregated KPIs, where the aggregation is done over several shared attributes such as geographical region and data center~\cite{squeeze2019, gu2020mid}. Each attribute has a range of categorical values on which to aggregate, and the data can thus be seen as multidimensional time series data. We refer to the attribute combinations and their attribute values at different levels of aggregation as elements.
When anomalous behavior is identified in an aggregated measure, we need to quickly locate where the fault lies, that is, identify the root cause. Specifically, we want to localize the set of elements that best explains the anomaly. 

\begin{table}[t]
    \centering
    \small
    \caption{Example of a multi-dimensional measure with two attributes. The abnormal rows are in bold and the best root cause set is $\{(X, *)\}$.}
    \label{tab:fundametal}
    \begin{tabular}{cccc}
         \toprule
         Data Center & Device Type & Actual & Forecast \\
         \midrule
         \textbf{X} & \textbf{D1} & \textbf{10} & \textbf{30} \\
         \textbf{X} & \textbf{D2} & \textbf{3}  & \textbf{10} \\
         Y & D1 & 15 & 14 \\
         Y & D2 & 30 & 30 \\
         Y & D3 & 100 & 102 \\
         \midrule
         \multicolumn{2}{c}{Total} & 158 & 186 \\
         \bottomrule
    \end{tabular}
\end{table}

To identify the root cause of an anomaly, the expected (forecasted) values for each element are compared to the actual values in the abnormal period. 
Table~\ref{tab:fundametal} shows an example with two attributes where the aggregated measure (total) is abnormal with the root cause being $\{(X, *)\}$, where $*$ indicates an aggregated attribute. Due to the nature of the problem, the root cause is a set of elements with different levels of aggregation. The main challenge is the huge search space since we need to consider all possible combinations of any number of attribute values. 
Even for tiny toy examples such as the one in Table~\ref{tab:fundametal}, the number of potential root causes is $2^{2+3+5}-1$. 
For a measure with $d$ dimensions each with $n$ values, the number of valid elements are $\sum_{i=1}^d {d \choose i}n^i=(n+1)^d-1$ which gives $2^{(n+1)^d-1}-1$ number of possible combinations. 

Evidently, it is not possible to examine the whole search space for even reasonably small $d$ and $n$, hence, previous works have focused on reducing the search space. Adtributor~\cite{adtributor2014} and iDice~\cite{idice2016} focus only on simpler cases. Adtributor can only identify root causes that lie within a single attribute, while iDice assumes a low number of root cause elements. R-Adtributor~\cite{masterthesis} is a recursive version of Adtributor that incorporates the entire search space; however, the stop condition for the recursion is hard to adjust correctly. Apriori~\cite{apriori2017} suffers from the same tuning difficulty, and its running time is an order of magnitude slower than the rest.
HotSpot~\cite{hotspot2018} can be slow and has problems finding root causes with many elements with low levels of aggregation. Squeeze~\cite{squeeze2019} proposes a clustering approach and top-down search for each cluster, however, it is highly sensitive to the clustering outcome and has difficulty with very fine-grained anomalies. AutoRoot~\cite{autoroot2021} improves on Squeeze, however, it cannot identify multiple elements in the same root cause and is still highly dependent on clustering accuracy.

To resolve the above challenges, we propose \textit{RiskLoc}, which partitions the data into two sets, normal and anomalous, and iteratively identifies high-risk root cause elements by applying our proposed \textit{risk score}. The risk score consists of two components, $r_1$ and $r_2$, where $r_1$ considers the weighted proportion of an element in the normal and anomalous partitions while $r_2$ compares the impact when adjusting for the \textit{ripple effect}~\cite{hotspot2018} (a property stating that changes in an element will propagate to its descendant elements). Iteration stops when the identified elements explain the abnormal divergence to a satisfactory degree. 
Previous works have mainly used proprietary data for method evaluation, and only \cite{squeeze2019} have made various semi-synthetic datasets publicly available. Unfortunately, there is a distinct lack of real-world datasets, and the semi-synthetic ones are of relatively small-scale. To demonstrate the effectiveness of our approach, we introduce multiple synthetic datasets with different characteristics, including one that is multiple magnitudes larger than previous datasets.

Our main contributions are as follows.
\begin{itemize}
    \item We propose RiskLoc to accurately and efficiently locate root causes in multi-dimensional time series without any significant root cause assumptions. We exploit a partitioning scheme and iteratively search for probable root cause elements within the abnormal partition. A heuristic risk score is proposed to appraise elements by integrating two key components: 1) the weighted deviation score distribution of an element and 2) the impact of adjusting for the ripple effect property. 
    \item We introduce three fully synthetic benchmark datasets for comprehensive algorithm evaluation that complement the publicly available semi-synthetic datasets~\cite{squeeze2019} since these datasets do not cover the entire search space; there are no root causes with multiple elements in the same cuboid and the datasets are all relatively small-scale. Furthermore, the code for the synthetic dataset generation is released, together with code for RiskLoc and baselines, to help alleviate the problem of data scarcity in the field.\footnote{Source code is available at \url{https://github.com/shaido987/riskloc}.}
    \item We empirically demonstrate the advantages of RiskLoc with extensive experiments on multiple datasets. RiskLoc consistently outperforms state-of-the-art baselines. On the most challenging semi-synthetic dataset, RiskLoc attains an improvement of 57\% over the second-best approach, while the gains on the synthetic datasets is up to 41\%.
\end{itemize}



%



%

%
\section{Related Work}
\label{sec:related}


Root cause localization in multidimensional measures is becoming increasingly relevant with increasing system size and higher demand for service reliability. Several previous works have focused on specific scenarios (e.g., advertising systems~\cite{adtributor2014, masterthesis}) and, more recently, generic approaches~\cite{squeeze2019, autoroot2021}. The pioneering work in the area is Adtributor~\cite{adtributor2014} which uses explanatory power and surprise metrics to identify the set of most likely root causes; however, it is limited to finding root causes in a single attribute. R-Adtributor~\cite{masterthesis} extends Adtributor to handle root causes in multiple attributes by recursively running Adtributor. However, the termination condition of when to stop the recursion is hard to specify, often returning inexact and verbose root causes sets. 

Apriori~\cite{apriori2017} was proposed as part of a system for detection and localization of end-to-end performance degradations at cellular service providers. It identifies leaf elements where the actual values deviate from the predicted values and then applies association rule mining techniques to localize the root cause. The method is sensitive to the parameter selections for support and confidence, requiring manual fine-tuning depending on the data. Furthermore, the running time is often too long for real-time usage. 

iDice~\cite{idice2016} uses various pruning strategies to identify effective attribute combinations (referred to as elements in this work) of emerging issues within issue reports. 
However, it is created to handle simpler cases where only a few of the elements are affected by the anomaly and will thus perform poorly in more complex situations.

Hotspot~\cite{hotspot2018} adopts Monte Carlo tree search (MCTS) to manage the large search space. To evaluate the element sets, they introduce the ripple effect property and, building on this, a potential score for ranking potential root causes. HotSpot has difficulty identifying more complex root causes involving elements with many attributes or multiple elements. 
Moreover, both HotSpot and iDice focus on simple (fundamental) measures and are not designed to handle nonadditive measures.

As established by \cite{squeeze2019}, the above works may miss elements that have insignificant anomaly magnitudes, and they propose Squeeze to rectify this issue. To prune the search space, a density-based clustering on the leaf elements relative residuals is used and each cluster is searched for a root cause set by leveraging a generalized potential score. The assumption is made that each cluster only contains elements with the same root cause, which may not always be true. The final result is highly dependent on the clustering accuracy, which involves hard-to-determine parameters. In addition, Squeeze relies on a heuristic method to sort the elements before scoring the partitions, however, this ranking fails for the most fine-grained elements and can result in numerous false positives.

More recently, AutoRoot~\cite{autoroot2021} improves upon Squeeze by introducing parameter-free clustering, relative scoring, and simplified filtering strategy. However, the issues of reliance on accurate clustering and the assumption of a single root cause per cluster are still present. AutoRoot makes further assumptions on the root cause and removes the ability to return multiple elements from the same cluster to speed up execution.


\section{Preliminaries}
In this section, we begin by defining the problem, the related notation, and terminology. Then, we briefly introduce the ripple effect property, a key assumption for the success of the algorithm. Finally, we give a simplified example to clarify the problem.

\subsection{Problem Formulation and Terminology}
When an anomaly is identified in an aggregated measure, we want to localize the combination of attributes and attribute values that best explain the anomalous behaviour. We use time series data or log data with timestamps as input and aggregate all values with the same accompanying attributes within the anomaly interval; e.g., in Table~\ref{tab:fundametal}, the time information has been removed, and each attribute combination has a single aggregated value.

More formally, we consider an aggregated measure $M$ with a set of attributes $\mathcal{A} = \{A_1, A_2,...,A_d\}$ where each attribute $A_i \in \mathcal{A}$ contains a set of $n_i$ distinct categorical values $\mathcal{V}_i = \{v_1, v_2,..., v_{n_i}\}$. We denote a distinct combination of attribute values $\{(s_1,...,s_d) | s_1 \in \{\mathcal{V}_1 \cup *\}, ..., s_d \in \{\mathcal{V}_d \cup *\} \}$ as an \textit{element}, where a wildcard~$*$ denotes aggregation in an attribute following the notation in~\cite{hotspot2018}. 
For example, for a measure with $d=4$ attributes, we can describe the fully aggregated measure as $(*,*,*,*)$. Similarly, a set of elements $\{(s,*,*,*) | s \in \mathcal{V}_1 \}$ have their second, third, and fourth attributes fully aggregated. 
Our goal is to localize the root cause set $RS$ that contains the elements, at any level of aggregation, that together offer the best explanation of the anomalous behaviour while being as succinct as possible.

The most fine-grained elements are denoted as \textit{leaf elements}. These elements do not have any aggregation and the set of leaf elements is thus denoted as $\mathcal{E} = \{(s_1,s_2,s_3,s_4) | s_1 \in \mathcal{V}_1, s_2 \in \mathcal{V}_2, s_3 \in \mathcal{V}_3, s_4 \in \mathcal{V}_4 \}$. An element $e_2=(u_1,..,u_d)$ is considered a proper descendant of $e_1=(v_1,...,v_d)$ iff
\begin{equation}
    (e_1 \neq e_2) \wedge \bigwedge_{i=1}^{d} (u_i = v_i \vee v_i = *).
\end{equation}
For convenience, we define the set of descendant leaf elements of an element $e$ as:
\begin{equation}
LD(e) = \{ e' | (e'\text{ descendant of } e) \wedge (e' \text{ is a leaf element}) \}.
\end{equation}

\begin{figure}[t]
    \centering
    \includegraphics[trim={0.5cm 0.3cm 0.4cm 0.5cm}, clip, width=0.48\textwidth]{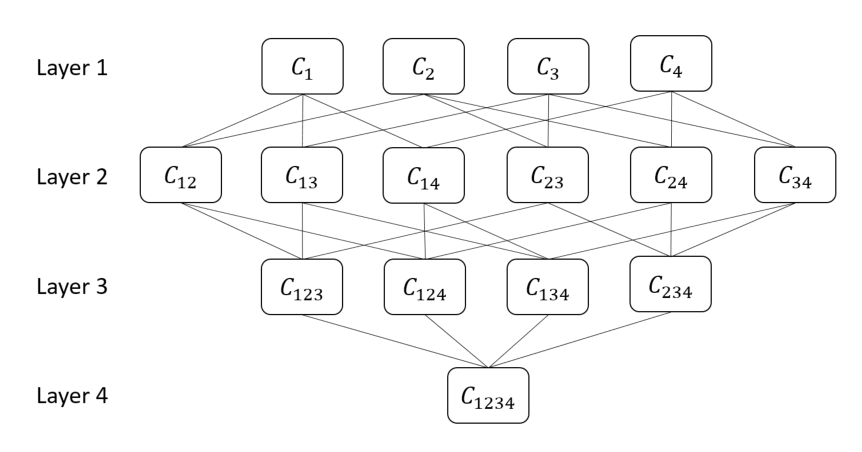}
    \caption{Cuboid relation graph with $d=4$.}
    \label{fig:cuboid}
\end{figure}

Based on the different degrees of aggregation, we form sets of elements called \textit{cuboids}~\cite{datamining}. We denote the set of all cuboids as $\mathcal{C}$ where a single cuboid $C_k \in \mathcal{C}$ ($k$ being an arbitrary combination of ${1,...,d}$) consists of the elements whose aggregated attributes are the same. The cuboid containing leaf elements is thus denoted as $C_{1234}$.
Cuboids can be separated into different \textit{layers} determined by the number of fully aggregated attributes, as shown in Figure~\ref{fig:cuboid} where the lines represent the parent-child relationship between the cuboids.
We further denote $ES(C)$ to be the element set of cuboid $C$, e.g., we have $ES(C_1) = \{(s,*,*,*) | s \in \mathcal{V}_1 \}$.

Two types of measures are considered, \textit{fundamental} and \textit{derived} measures~\cite{adtributor2014}. Fundamental measures are additive and can easily be divided into segments following their attribute values. Page views, number of items, and response times are examples of fundamental measures. On the other hand, derived measures are obtained by applying a function on fundamental measures and are generally nonadditive. 
Relevant examples are revenue per page view and success rates, i.e., quotients of fundamental measures. Derived measures may require more complex handling depending on the applied method~\cite{squeeze2019}.

Each element will have a \textit{actual} value and a \textit{forecast} value. We denote these as $v(\cdot)$ and $f(\cdot)$, respectively. The actual value is measured in the monitored system while the forecast value is the result of a forecasting algorithm, e.g., Prophet~\cite{taylor2018forecasting}, ARMA~\cite{pincombe2005anomaly}, or simple moving average. 
For fundamental measures, the values of the leaf elements can be found directly while the values of the non-leaf elements are simply the sum of its descendant leaf elements, i.e.,
\begin{align}
    & v(e) = \sum_{e' \in LD(e)} v(e'), & f(e) = \sum_{e' \in LD(e)} f(e').
\end{align}

For a derived measure $M_d$, we denote the function to create the derived measure as $h(\cdot)$, where $M_d = h(M_1,...,M_n)$ and $\{M_1,...,M_n\}$ are the fundamental measures used. For non-leaf elements, the aggregated actual values are computed as $v(e)=h(v_1(e),...,v_n(e))$, where $\{v_1(e),...,v_n(e)\}$ are the actual values of the respective fundamental measure. The forecast values are computed analogously as $f(e)=h(f_1(e),...,f_n(e))$.

Another relevant term is the \textit{explanatory power}~\cite{adtributor2014} of an element which is defined as the portion of change in the overall measure that can be explained by the change in the given element's value. For an element $e$ of a fundamental measure $M$, we have:
\begin{equation}
\label{eq:explanatory_power}
    ep(e) := \frac{v(e) - f(e)}{v(M) - f(M)}.
\end{equation}
For derived measures, we refer to the derivation given in~\cite{adtributor2014}.

\subsection{Ripple Effect and Deviation Score}
The \textit{ripple effect} was introduced by~\citet{hotspot2018} and was further generalized to derived measures and zero forecast values by~\citet{squeeze2019}. The property characterizes how a change in an element is propagated to its descendant leaf elements. Since the actual and forecast values of an element $e$ are the aggregation of its descendant leaf element values, a change in $e$ will undoubtedly affect the elements in $LD(e)$. The ripple effect property states that an element $e_r$ in the root cause set $RS$ will affect all elements $e \in LD(e_r)$ following the proportions of their forecast values, that is:
\begin{equation}
    \frac{f(e) - v(e)}{f(e)} = \frac{f(e_r) - v(e_r)}{f(e_r)},\ (f(e_r) \neq 0).
\end{equation}
The extension in~\cite{squeeze2019} to \textit{generalized ripple effect} (GRE) replaces $f(\cdot)$ with $\frac{f(\cdot)+v(\cdot)}{2}$ to allow for zero forecast values. The enhanced property is thus as follows:
\begin{equation}
\label{eq:gre}
    \frac{f(e) - v(e)}{f(e) + v(e)} = \frac{f(e_r) - v(e_r)}{f(e_r) + v(e_r)}.
\end{equation}
Following this property, the \textit{deviation score}~\cite{squeeze2019} of an element $e$ is defined as:
\begin{equation}
\label{eq:deviation_score}    
    ds(e) := 2 \cdot \frac{f(e) - v(e)}{f(e) + v(e)},
\end{equation}
which adhering to GRE will theoretically give all leaf elements in the same root cause similar scores.

\subsection{Problem Intuition}
To clarify the problem of localizing the set of elements that best explains the anomaly, we take the simple example in Table~\ref{tab:fundametal}. 
We denote the two cuboids in the first layer as $C_c$ and $C_t$, representing the data center and device type, respectively. The second layer has a single cuboid $C_{ct}$ with leaf elements.
Each cuboid contains a set of elements with the relevant aggregations, i.e., $E(C_c) = \{ (X, *), (Y, *) \}$, $E(C_t) =\{ (*, D1), (*, D2), (*, D3) \}$, and $E(C_{ct}) = \{ (X, D1), (X, D2), (Y, D1), (Y, D2), (Y, D3) \}$.

The anomaly detection is performed on the fully aggregated measure; in Table~\ref{tab:fundametal} the aggregated forecast value is $f(*,*)=186$, while the actual value is $v(*,*)=158$. The best set of each cuboid are compared to determine the best overall element set.
For cuboid $C_t$, we can determine that $(*, D3)$ is likely not affected by the anomaly. For $(*, D1)$ and $(*, D2)$ the aggregated values are $44$ and $40$ while the actual values are $25$ and $33$, making this a possible root cause. However, in $C_c$, $(X, *)$ offers a better and more succinct explanation of the root cause, as it contains only the two abnormal leaf elements. Succinctness (i.e., Occam's razor principle) is also the reason to prefer $(X, *)$ in favor of the root cause set in $C_{ct}$: $\{(X, D1), (X, D2) \}$. The set of elements that best explain the root cause is therefore $\{(X, *)\}$.

\section{Methodology}

\begin{figure}[t]
    \centering
    \includegraphics[trim={0cm 0cm 0cm 0cm}, clip, width=1.0\linewidth]{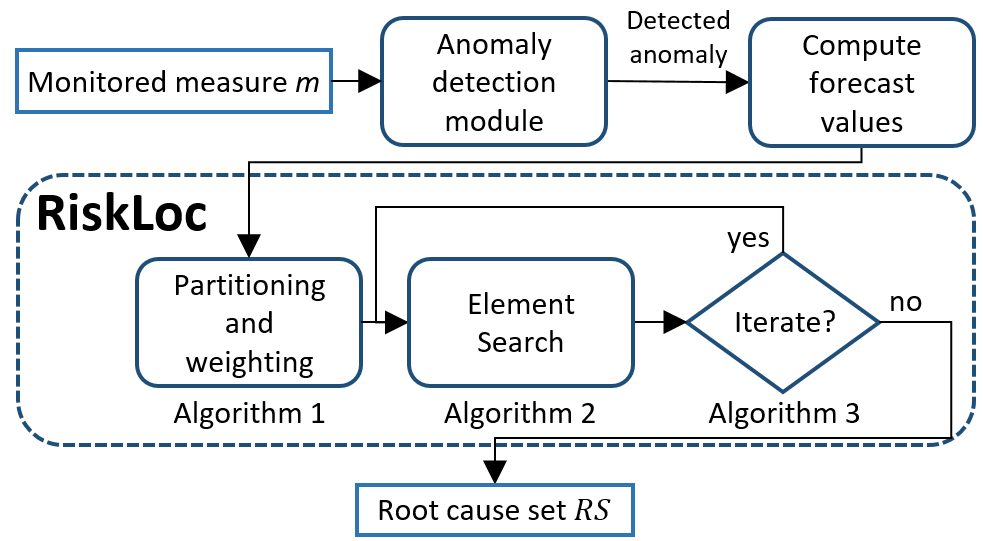}
    \caption{An overview of RiskLoc.}
    \label{fig:overview}
\end{figure}

In this section, we introduce RiskLoc, our proposed solution to the multi-dimensional root cause localization problem. An overview of RiskLoc is shown in Figure~\ref{fig:overview}.
In essence, when an anomaly is detected by some anomaly detection module on a monitored measure, RiskLoc is initiated and the search for the root cause location is started. The actual values of all leaf elements are acquired and aggregated for the anomaly time interval and the forecast values are computed from historical data. 
RiskLoc employs three main components in its search for the root cause set, 1) leaf element partitioning and weighting, 2) risk score, and 3) element search and iteration. The details of each component and an optional element pruning strategy are presented below.

\begin{figure}[t]
    \centering
    \includegraphics[trim={0.1cm 0.75cm 2.75cm 2.25cm}, clip, width=0.475\textwidth]{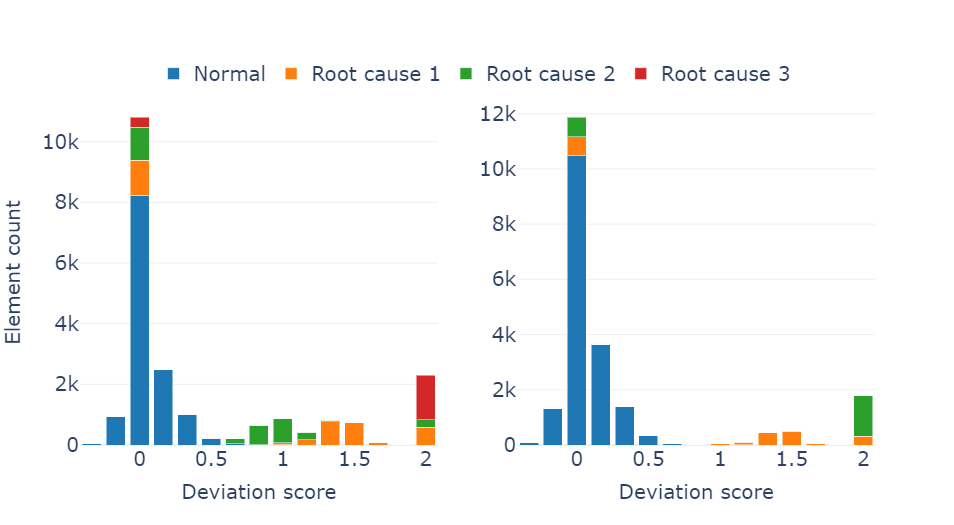}
    \caption{Two examples of typical deviation score distributions.}
    \label{fig:deviation-scores}
\end{figure}

\subsection{Leaf Element Partitioning and Weighting}
We compute the deviation score for each leaf element using Equation~\ref{eq:deviation_score}. Leaf elements with small deviation scores can be considered normal, as their relative forecast residuals are small, while deviation scores further from zero may be anomalous. Two histograms illustrating typical deviation score distributions can be seen in Figure~\ref{fig:deviation-scores}. Note that the deviation scores of the anomalous leaf elements share the same sign. This is generally the case as an anomaly was detected in the aggregated measure, anomalies with conflicting sign risk cancel each other when aggregating and are thus less likely to be detected.

\renewcommand{\algorithmicrequire}{ \textbf{Input:}} 
\renewcommand{\algorithmicensure}{ \textbf{Output:}}
\algnewcommand\algorithmicreturn{\textbf{return}}
\algnewcommand\RETURN{\State \algorithmicreturn}
\newcommand{\algorithmicbreak}{\textbf{break}}
\newcommand{\BREAK}{\STATE \algorithmicbreak}

\algnewcommand{\IfThenElse}[3]{
  \State \algorithmicif\ #1\ \algorithmicthen\ #2\ \algorithmicelse\ #3}

\begin{algorithm}[t]
	\caption{Partitioning and Weighting}
	\label{alg:weight_function}
	\begin{algorithmic}[1]
	    \REQUIRE leaf element set $\mathcal{E}$
		\STATE $\mathcal{D} = \{ds(e) \mid e \in \mathcal{E} \}$
		\STATE Remove outliers in $\mathcal{D}$
		\IF{$\left|\min\mathcal{D}\right| < \left|\max\mathcal{D}\right|$}
		    \STATE $t = -\min\mathcal{D}$
		    \STATE $E_n = \{ (e, \left| t - ds(e) \right|, 0) \mid e \in \mathcal{E}, ds(e) < t,$ \\ \phantom . \phantom . \phantom . \phantom . \phantom . \phantom . \phantom . \phantom .  $f(e) \neq 0 \vee v(e) \neq 0 \}$
		    \STATE $E_a = \{ (e, \left| ds(e) \right|, 1) \mid e \in \mathcal{E}, ds(e) \geq t \}$
		\ELSE
		    \STATE $t = -\max\mathcal{D}$
		    \STATE $E_n = \{ (e, \left| t - ds(e) \right|, 0) \mid e \in \mathcal{E}, ds(e) > t,$ \\ \phantom . \phantom . \phantom . \phantom . \phantom . \phantom . \phantom . \phantom . $f(e) \neq 0 \vee v(e) \neq 0 \}$
		    \STATE $E_a = \{ (e, \left| ds(e) \right|, 1) \mid e \in \mathcal{E}, ds(e) \leq t \}$
		\ENDIF
		\STATE $E_z = \{ (e, 0, 0) \mid e \in \mathcal{E}, f(e) = 0,  v(e) = 0 \}$
		\STATE $\mathcal{E}_w = E_n \cup E_a \cup E_z$ 
		\STATE $\mathcal{E}_w = \{(e, \min\{w, 1\}, p) \mid (e,w,p) \in \mathcal{E}_w \}$
		
    \ENSURE Weighed leaf element set $\mathcal{E}_w$
	\end{algorithmic}
\end{algorithm}

We use deviation scores to separate the leaf elements into a normal and an abnormal set using a simple 2-way partitioning scheme. The elements are given weights corresponding to their distance from the partitioning point. The complete procedure is detailed in Algorithm~\ref{alg:weight_function}. 
A suitable partitioning point is determined by relying on the simple assumption that normal elements are relatively evenly distributed around zero, i.e., that the forecasts are reasonably accurate, and that the anomalous leaf element share the same sign. 
We capture the maximum absolute deviation score (adjusting for a few outliers; we simply remove the top and bottom $5$ unique values) in the opposite direction to the anomaly and flip the sign. This estimates the maximum absolute deviation score of the normal data. The anomaly direction is simultaneously inferred from the data but could alternatively be obtained from the anomaly detection procedure.

Each leaf element is given a weight corresponding to the distance from the partitioning point. For the anomalous elements, the weight is the deviation score (i.e., a linear increase from the partitioning point) with a maximum value of 1. For the normal leaf elements, we similarly increase the weight linearly from the partitioning point; however, we start the weights with 0. In this way, normal leaf elements close to the partitioning point will have small weights. The influence of these leaf elements is thus reduced since they are more likely to be anomalous than points further away. Therefore, the accuracy of the partitioning scheme will have a lower impact on the final result, in contrast to the rigid clustering used by Squeeze and AutoRoot. 
We further note that leaf elements with both $v(e)=0$ and $f(e)=0$ do not contain any information to determine whether their parent elements
are anomalous or not. The weights of these leaf elements are thus set to 0.

\begin{algorithm}[t]
	\caption{Element Search (ES)}
	\label{alg:element_search}
	\begin{algorithmic}[1]
		\REQUIRE weighed leaf element set $\mathcal{E}_w$, 
		risk threshold $t_r$, explanatory power threshold $t_{ep}$
		
	    \FOR{$l=1$ {\bfseries to} $d$}
	        \STATE $candidates = \{\}$
	        \FOR{$C \in \mathcal{C}$ in layer $l$}
	            \STATE prune elements in $ES(C)$
	            \FOR{$e_r \in ES(C)$}
	            
		            \STATE $risk = \text{apply equation~\ref{eq:risk} on } e_r$
		            
		            \IF{$risk \geq t_r$ \textbf{and} $ep(e_r) \geq t_{ep}$}
		                \STATE $candidates = candidates \cup \{e_r\}$
		            \ENDIF
	            \ENDFOR
	        \ENDFOR
	        \IF{$candidates$ is not empty}
	            \RETURN{ $e_r \in candidates$ with maximum $ep(e_r)$}
	        \ENDIF
	    \ENDFOR
    \RETURN{ null}
	\end{algorithmic}
\end{algorithm}

\begin{algorithm}[t]
	\caption{RiskLoc}
	\label{alg:main_alg}
	\begin{algorithmic}[1]
		\REQUIRE leaf element set $\mathcal{E}$, risk threshold $t_r$, 
		proportional explanatory power threshold $t_{pep}$
		
		\STATE Obtain $\mathcal{E}_w$ using Algorithm~\ref{alg:weight_function}
		\STATE $R = \{ e \mid (e, w, p) \in \mathcal{E}_w, p = 1 \}$
		
		\IF{$ep(R) < 0$}
		    \STATE $ep(\mathcal{E}_w) = -ep(\mathcal{E}_w)$ \COMMENT{Negate $ep$ for all elements}
		\ENDIF
		
		\STATE $t_{ep} = t_{pep} \cdot ep(R)$
		
		\STATE $RS = \{\}$
		\WHILE{$ep(R) \geq t_{ep}$}
		    
            \STATE $e_r = \text{ES}(\mathcal{E}_w, t_r, t_{ep})$
            \IF{$e_r$ is null}
                \BREAK
            \ENDIF
            \STATE $\mathcal{E}_w = \{ (e, w, p) \mid (e, w, p) \in \mathcal{E}_w, e \not\in LD(e_r) \}$
            \STATE $R = \{ e \mid (e, w, p) \in \mathcal{E}_w, p = 1\}$
            \STATE $RS = RS \cup \{e_r\}$
		\ENDWHILE
    \ENSURE Set of root causes $RS$
	\end{algorithmic}
\end{algorithm}

\subsection{Risk Score}
We propose a heuristic risk score to identify potential root cause elements in each cuboid, i.e., high-risk elements. The risk score consists of two terms, $r_1$ and $r_2$, which reflect the two key aspects of the root cause elements: descendant elements are mainly in the anomalous partition, and the forecast residuals follow the property of the ripple effect.

For an element $e_r$, independent of the level of aggregation, we obtain the weighted sum of its descendant leaf elements separated by the partitioning scheme. Let $\mathcal{E}_w$ be the weighted leaf element set obtained by Algorithm~\ref{alg:weight_function} and $S$ the descendant leaf elements of $e_r$, i.e., $S=\{ (w, p) \mid  (e, w, p) \in \mathcal{E}_w, e \in LD(e_r) \}$, then we have:
\begin{align}
    & w_a = \smashoperator[r]{\sum_{(w, 1) \in S}} w, & w_n = \smashoperator[r]{\sum_{(w, 0) \in S}} w.
\end{align}
We define $r_1$ as follows:
\begin{equation}
\label{eq:high_risk}
    r_1 = \frac{w_a}{w_n + w_a + 1}.
\end{equation}
The purpose of $r_1$ is to verify whether $e_r$ is relevant to the anomaly. A high $r_1$ score indicates that the majority of the descendant leaf elements are within the anomalous partition. Note that the significance of elements in the highest layer is limited by adding $1$ to the denominator.

According to the generalized ripple effect in Equation~\ref{eq:gre}, the expected value of descendant leaf elements $e \in LD(e_r)$ should be $a(e) = f(e)\frac{v(e_r)}{f(e_r)}$. The difference between $v(e)$ and $a(e)$ quantifies the degree to which $e$ follows the ripple effect paradigm. Different from GPS~\cite{squeeze2019} and NPS~\cite{autoroot2021}, we propose a simplified score $r_2$ to be used in conjunction with $r_1$. Since $r_1$ already incorporates the relative importance of $e_r$, we do not need $r_2$ to consider leaf elements $e \not\in LD(e_r)$. Instead, we compare the deviation scores of all leaf elements $e \in LD(e_r)$ before ($r_d$) and after ($r_n$) adjusting for the ripple effect:
\begin{align}
    & r_n = \sum_{e \in LD(e_r)}{2 \left|\frac{a(e)-v(e)}{a(e)+v(e)}\right|},
    & r_d = \sum_{e \in LD(e_r)}{\left|ds(e)\right|}.
\end{align}
$r_2$ then directly quantifies how much the average deviation score has changed:
\begin{equation}
    \label{eq:low_risk}
    r_2 = \frac{r_n}{r_d}.
\end{equation}
If $r_2$ is close to zero then $e_r$ adhere to the ripple effect and is a more likely root cause candidate, on the other hand, a large value indicates that the relative forecast residuals of the descendant leaf elements of $e_r$ has a high variance, i.e., $e_r$ is unlikely to belong to the root cause set.

Finally, we combine $r_1$ and $r_2$ together into a single \textit{risk} metric:
\begin{equation}
    \label{eq:risk}
    risk = r_1 - r_2.
\end{equation}
Following the above definitions, we have that $0 \leq r_1 < 1$ and $r_2 \geq 0$, hence, $risk < 1$. Note that for leaf elements, $r_2$ will always return $0$, so we will rely entirely on $r_1$ for these.

\begin{table*}[t]
    \centering
    \small
    \caption{Summary of the datasets.}
    \label{tab:datasets}
    \begin{tabular}{ccccccccc}
    \toprule
    \multirow{2}*{\textbf{Dataset}} & \multirow{2}*{$\mathbf{n}$} & \multirow{2}*{$\mathbf{d}$} & \multirow{2}*{\textbf{\#elements}} & \multirow{2}*{\textbf{Type}} &
    \multirow{2}*{\textbf{Residual}} & \multicolumn{3}{c}{\textbf{Anomaly}}\\
    \cmidrule{7-9}
    &  &  &  &  &  & \textbf{Maximum} & \textbf{Max elements} &\textbf{Mean significance} \\
    \midrule
    $\mathcal{A}$   & $16{,}684$ & $5$   & $15{,}324$     & Fundamental & $3.92\%$ & $3$  & $1$       & $0.28$ \\
    $\mathcal{A}_\star$   & $9{,}430$ & $5$   & $15{,}324$     & Fundamental & $3.92\%$ & $5$  & $1$       & $0.28$ \\
    $\mathcal{B}_0$ & $9\cdot100$ & $4$   & $21{,}600$     & Fundamental & $0.80\%$ & $3$   & $1$       & $0.06$ \\
    $\mathcal{B}_1$ & $9\cdot100$ & $4$   & $21{,}600$     & Fundamental & $3.19\%$ & $3$   & $1$       & $0.07$ \\
    $\mathcal{B}_2$ & $9\cdot100$ & $4$   & $21{,}600$     & Fundamental & $6.38\%$ & $3$   & $1$        & $0.10$ \\
    $\mathcal{B}_3$ & $9\cdot100$ & $4$   & $21{,}600$     & Fundamental & $9.55\%$ & $3$   & $1$       & $0.12$ \\
    $\mathcal{B}_4$ & $9\cdot100$ & $4$   & $21{,}600$     & Fundamental & $12.7\%$ & $3$   & $1$       & $0.14$ \\
    $\mathcal{D}$   & $9\cdot100$ & $4$   & $21{,}600$     & Derived     & $3.99\%$ & $3$   & $1$     & $0.03$ \\
    \midrule
    $\mathcal{S}$   & $1{,}000$ & $5$   & $48{,}000$     & Fundamental & $0.0\%-20.2\%$ & $3$   & $3$       & $0.09$ \\
    $\mathcal{L}$   & $1{,}000$ & $4$   & $36{,}000$     & Fundamental & $0.0\%-8.02\%$ & $5$   & $1$       & $0.00$ \\
    $\mathcal{H}$   & $100$ & $6$   & $24{,}000{,}000$     & Fundamental & $0.0\%-19.9\%$ & $3$   & $3$       & $0.04$ \\
    \bottomrule
    \end{tabular}
\end{table*}

\subsection{Element Search and Iteration}
We do not make any assumption on the number of elements in the root cause set or whether the elements are in the same cuboid or not.
Instead, we identify the single most likely element $e_r$, remove its leaf elements $LD(e_r)$ from $\mathcal{E}_w$, and then repeat the search for the next candidate. The details of the element search are in Algorithm~\ref{alg:element_search} and the iteration is carried out in Algorithm~\ref{alg:main_alg}.

The element search is done from lower layer cuboids to higher layer ones following the Occam's razor principle, allowing for an early exit if a suitable candidate is found. For an element to be considered suitable, we require its risk score to be above a threshold $t_r$ and its explanatory power to be above a minimum threshold $t_{ep}$ (see Equation~\ref{eq:explanatory_power}). 
When multiple candidates from the same layer satisfy the conditions, we return the element with the highest explanatory power. 
To terminate the search and determine a suitable value for $t_{ep}$, we consider the total explanatory power of the leaf elements in the anomalous partition, see Algorithm~\ref{alg:main_alg}.
When the identified root cause set explains a sufficiently large proportion, the search is ended. This is determined by the parameter $t_{pep}$, where $0 < t_{pep} \leq 1$. We expect a proper root cause set to adequately explain the disparity between the actual and forecast values, which rationalizes our use of explanatory power, both as the termination condition and for selecting relevant elements. In this way, we ensure that each identified element is significant and explains at least $t_{pep}$ of the total anomaly. Note that we only need to adequately explain leaf elements in the anomalous partition, and anomalies with insignificant magnitudes~\cite{squeeze2019} can therefore still be identified. 

Depending on the direction of the anomaly, it may be necessary to adjust the sign of the elements' explanatory power to ensure that their values are positive. In Algorithm~\ref{alg:main_alg} (lines 3 to 5), we consequently examine the total sum of explanatory power in the anomalous partition; if it is negative, then we switch the sign for all elements. 

\subsection{Element Pruning}
An optional pruning of the elements can be done to increase the efficiency of the algorithm and decrease the running time, without having any impact on the returned result. See line 4 in Algorithm~\ref{alg:element_search}. We identify any element $e$ whose leaf elements set $LD(e)$ has a maximum potential explanatory power less than $t_{ep}$ and remove these and all of their descendant elements from the search space. We compute the maximum potential explanatory power of an element $e$ considering the optimal subset of descendant elements:
\begin{equation}
    \label{eq:element_prune}
    ep^+(e) = \smashoperator[r]{\sum_{e' \in LD(e)}} \max\{0, ep(e')\}.
\end{equation}
If $ep^+(e) < t_{ep}$, then no subset of descendant elements of $e$ will have an explanatory power high enough to be selected and can safely be pruned. The benefit of element pruning decreases at higher layers since there are fewer descendant elements, and we can therefore limit the element pruning to the first few layers. 

\section{Evaluation}
In this section we begin by introducing the experimental setup, delineating the synthetic dataset generation, datasets, and evaluation metrics. This is followed by a comparison between RiskLoc and related work, a parameter sensitivity analysis, and an ablation study.

\subsection{Experimental Setup}
\label{sec:experimental_setup}
\subsubsection{Synthetic Dataset Generation}
We generate synthetic data that emulates real-world data and complements the available semi-synthetic datasets~\cite{squeeze2019}. To generate these synthetic data instances, instead of generating time series, we directly construct the summed-up values for each separate element to simplify the generation process. As such, there is no concept of time involved and each leaf element $e$ only has a single actual value $v(e)$ and a single forecast value $f(e)$. 
Actual values $v(e)$ are sampled from a one-parameter Weibull distribution~\cite{Weibull1951ASD} with $\alpha \sim U[0.5,1.0]$, where $U$ is the uniform distribution with a closed interval. This is a subexponential, heavy-tailed distribution with a single tail, allowing the elements to take a broad and diverse range of values, with the majority being relatively closer to $0$. For aesthetic purposes, we multiply all values by $100$. We then set the actual values of an element to $0$ with a probability $p \sim U[0,0.25]$ to increase the difficulty in localizing root causes and to mirror a characteristic commonly seen in real world data.

For normal leaf elements (i.e., not in the root cause), we emulate the forecast residuals by adding Gaussian noise to the actual values with some value $\sigma$: 
\begin{equation}
    \label{eq:synth_forecast}
    f(e) = v(e) \cdot N(1, \sigma)
\end{equation}

For each generated instance, the number of anomalies and the number of elements involved in each is randomized from pre-set ranges. Then, each anomaly is randomly assigned a severity level and a deviation, corresponding to its distance from the normal samples and the variance of its leaf elements. Note that there is no guarantee that the root causes are significant. The full details of each generated data set can be found in Appendix~\ref{app:synthetic}. 

\begin{figure*}[t]
    \centering
    \includegraphics[trim={0.0cm 3.5cm 2.5cm 2.5cm}, clip, width=1.0\textwidth]{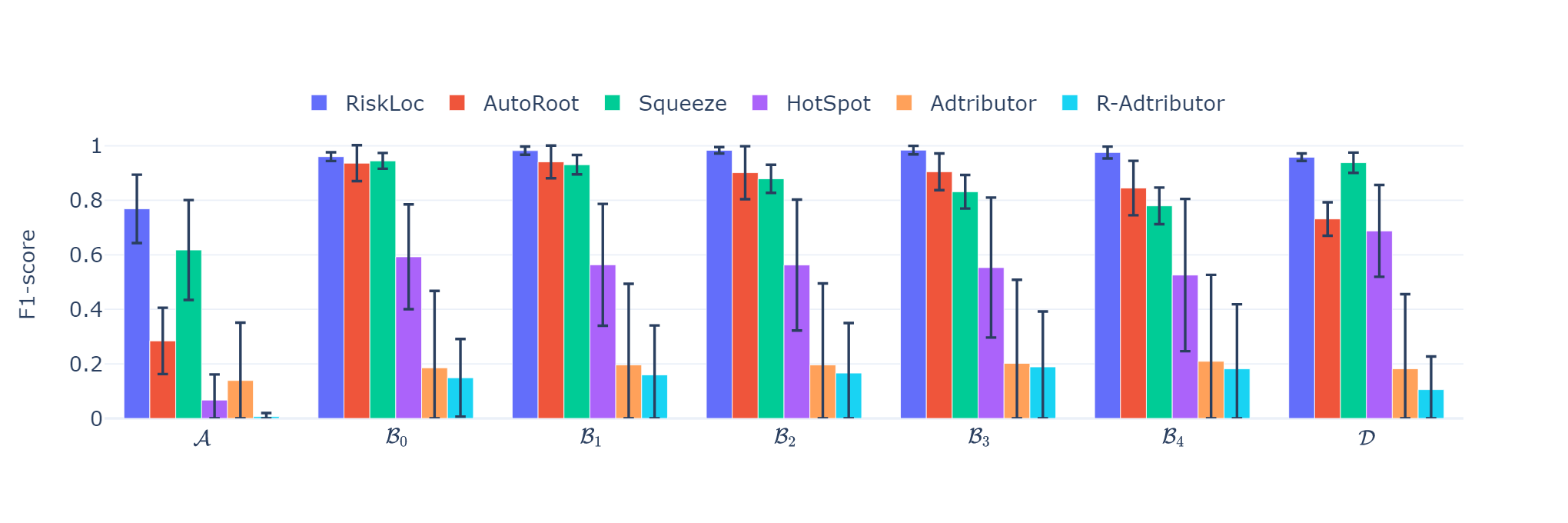}
    \caption{F1-score comparison for the simple datasets. 
    }
    \label{fig:f1_scores}
\end{figure*}

\subsubsection{Datasets}
We use the semi-synthetic datasets from~\citet{squeeze2019} for evaluation and further introduce three entirely synthetic datasets, $\mathcal{S}$, $\mathcal{L}$, and $\mathcal{H}$, which are generated using the approach described above. See Table~\ref{tab:datasets} for a summary of all datasets. 

$\mathcal{A}$, $\mathcal{B}_i$ with $i \in [0,..,4]$, and $\mathcal{D}$ are relatively simple; Each dataset has $9$ scenarios, with $100$ instances for each scenario. The difference between the scenarios is the number of elements in the root cause and in which layer the root cause is inside, both with values between $1$ and $3$. 
$\mathcal{D}$ have a structure similar to $\mathcal{B_i}$ but contain a derived measure: a quotient of two fundamental measures with a normal success rate following $U[0.9,1.0]$, see~\cite{squeeze2019} for the full details.
$\mathcal{A}_\star$ is a more challenging dataset, we divide $\mathcal{A}$ as obtained from \citet{squeeze2019} into $\mathcal{A}$ and $\mathcal{A}_\star$, where instances with root causes in layers lower than $3$ or with more than $3$ elements are put into $\mathcal{A}_\star$ and the others in $\mathcal{A}$. Note that our $\mathcal{A}$ corresponds to $\mathcal{A}$ as used for evaluation in~\cite{squeeze2019}.

The root causes in the semi-synthetic datasets do not cover the full search space, as only one element is allowed in each anomaly (albeit with up to three simultaneous anomalies in a single instance, but these must be from different cuboids), and all anomalies are always in the same layer. Moreover, the maximum forecast residual in the normal leaf elements is only around 13\%. 
In our synthetic datasets, we allow for higher forecast residuals, multiple root cause elements from the same cuboid, and from different layers within the same instance. Moreover, in contrast to~\citet{squeeze2019}, the anomaly direction is randomized (but the same for all anomalies in a single instance). Overall, the synthetic datasets are intended to be more complex and difficult.

The datasets $\mathcal{S}$ and $\mathcal{H}$ allow multiple elements in each anomaly (i.e., within the same cuboid) and can have up to $9$ separate elements in the root cause. On the other hand, $\mathcal{L}$ only has anomalies in the highest layer where none of the root cause elements have been aggregated in any dimension. These are typically insignificant compared to the numerous normal elements and are thus difficult to correctly identify. $\mathcal{H}$ is multiple orders of magnitudes larger than the other datasets with $24$ million leaf elements. In real-world scenarios, dimensions and attribute combinations of this size is not uncommon and $\mathcal{H}$ is created to verify the algorithms' effectiveness on these larger search spaces.

\subsubsection{Evaluation Metrics}
We assess the effectiveness of the methods using the F1-score. The F1 score is calculated at the element level, where a correctly reported element is considered a true positive (TP), a missed element is a false negative (FN), and any additional identified elements are false positives (FP). The F1-score is computed from the sum of all TP, FN, and FP as:

\begin{equation}
    \text{F1-score} = \frac{2 \cdot TP}{2 \cdot TP + FP + FN}.
\end{equation}
Another key metric is the efficiency of the algorithms. We consequently evaluate the average running time of the algorithms in different scenarios. 

\begin{table*}[t]
    \centering
    \small
    \caption{F1-scores on the more complex datasets.}
    \label{tab:hard_results}
    \begin{tabular}{ccccccc}
         \toprule
         \multirow{2}*{\textbf{Dataset}} & \multicolumn{6}{c}{\textbf{Algorithm}} \\
         \cmidrule{2-7}
         & \textbf{Adtributor} & \textbf{R-Adtributor} & \textbf{HotSpot} & \textbf{Squeeze} & \textbf{AutoRoot} & \textbf{RiskLoc} \\
         \midrule
         $\mathcal{A}_\star$ & $0.0650 \pm 0.12$ & $0.00003 \pm 0.00$ & $0.0127 \pm 0.02$ & $0.2951 \pm 0.16$ & $0.1802 \pm 0.06$ & $\mathbf{0.4640 \pm 0.10}$ \\
         $\mathcal{S}$ & $0.0589$ & $0.0066$ & $0.1740$ & $0.1283$ & $0.4478$ & $\mathbf{0.6350}$ \\
         $\mathcal{L}$ & $0.0000$ & $0.0089$ & $0.1848$ & $0.3524$ & $0.5266$ & $\mathbf{0.6767}$ \\
         $\mathcal{H}$ & $0.0235$ & $0.0000$ & $0.1109$ & $0.0511$ & $0.3492$ & $\mathbf{0.4906}$ \\
         \bottomrule
    \end{tabular}
\end{table*}

\subsection{Experimental Results}
We evaluate our proposed method against the following previous works: Adtributor~\cite{adtributor2014}, R-Adtributor~\cite{masterthesis}, HotSpot~\cite{hotspot2018}, Squeeze~\cite{squeeze2019}, and AutoRoot~\cite{autoroot2021}. Additional information regarding the running environment and parameter settings can be found in Appendix~\ref{app:exp_details}.

\subsubsection{Effectiveness}
The average F1-scores of each algorithm on the relatively simple datasets are presented in Figure~\ref{fig:f1_scores} together with the standard deviation over each root cause setting. See Appendix~\ref{app:add_exp_res} for the full table. 
As can be observed in the figure, RiskLoc consistently outperforms all baselines on all datasets with comparatively small performance deviations between different root cause settings within each dataset. We note a decline in the results for all baselines on the $\mathcal{B}_i$ datasets as the residual of the normal elements gets higher, while RiskLoc has more stable performance. 
In the dataset $\mathcal{A}$, there is significant overlap between anomalies that appear to severely affect AutoRoot at the lower layers, resulting in markedly worse performance compared to RiskLoc and Squeeze.
Notably, R-Adtributor has worse overall results than Adtributor, despite being able to handle deeper anomalies. This can be explained by Adtibutor performing considerably better on first-layer anomalies, regardless of the number of elements in the root cause.

The results on the more complex datasets are presented in Table~\ref{tab:hard_results}. RiskLoc achieves the best performance on all datasets with significant improvements over the second-best algorithms. For the most difficult semi-synthetic dataset, $\mathcal{A}_\star$, the relative increase in F1-score is 57\% and on the synthetic $\mathcal{S}$ dataset, the gain is 41\%. The results on $\mathcal{L}$ illustrate how Adtributor is unable to find fine-grained root causes, while R-Adtributor and HotSpot are of limited use. Squeeze, AutoRoot, and RiskLoc are better able to handle these fine-grained root causes, with RiskLoc obtaining the highest F1-score with a gain of 28\% over Autoroot, the second best approach. Finally, on the large-scale dataset $\mathcal{H}$ used to verify the effectiveness of the algorithms in larger search spaces, RiskLoc's gain is 40\%.


\subsubsection{Efficiency}
\begin{figure}[t]
    \centering
    \includegraphics[trim={0.0cm 0.4cm 4.0cm 2.5cm}, clip, width=1.0\linewidth]{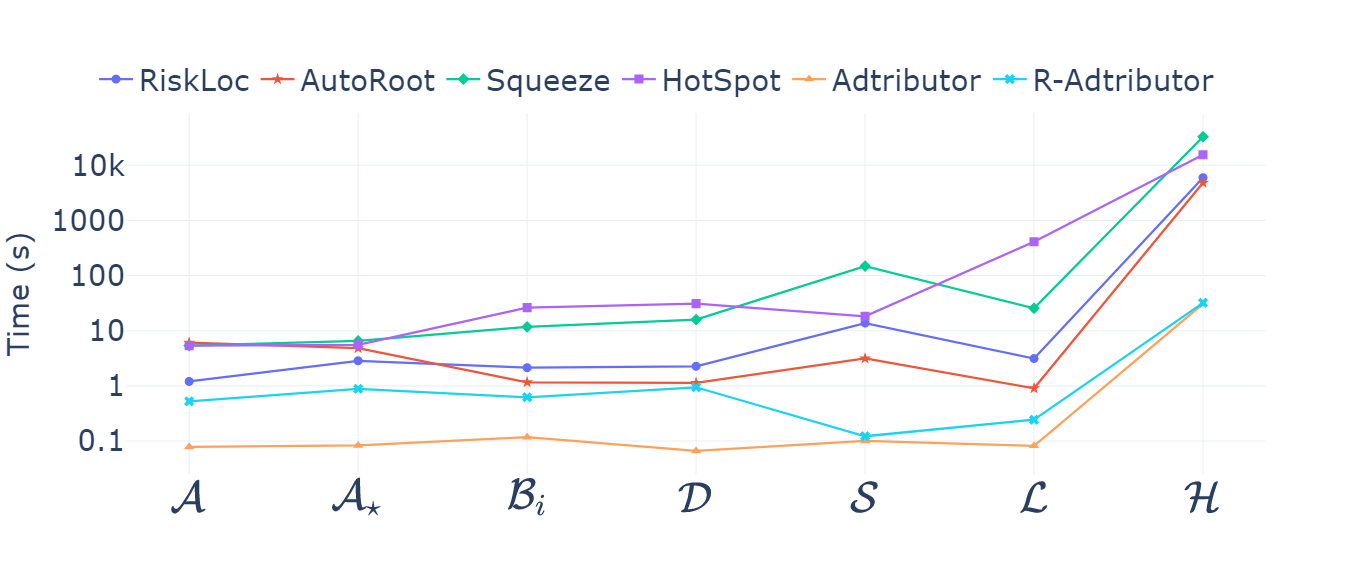}
    \caption{Average running time (s) on different datasets.}
    \label{fig:timing}
\end{figure}

The time consumption of each algorithm is presented in Figure~\ref{fig:timing}. 
As shown in the figure, Adtributor consistently consumes the least amount of time, while HotSpot and Squeeze are the slowest. 
RiskLoc runs in less than $3$ seconds on average on the simpler datasets, while $\mathcal{S}$ and $\mathcal{H}$ takes longer. 
The running time is clearly dependent on the search space size, while the root cause complexity has a smaller significance. 
Although the running time of Adtributor and R-Adtributor is significantly faster than that of RiskLoc, the accuracy is inadequate for all but the simplest root causes.
We further note that for most algorithms, including RiskLoc, there is a performance-speed trade-off, and in our parameter selection we have focused on the performance and the speed may thus have been negatively impacted. 

\subsection{Parameter Sensitivity Analysis}
\label{sec:param_sens}

\begin{figure}[t]
    \includegraphics[trim={0.0cm 0.0cm 2.0cm 2.0cm}, clip, width=1.0\linewidth]{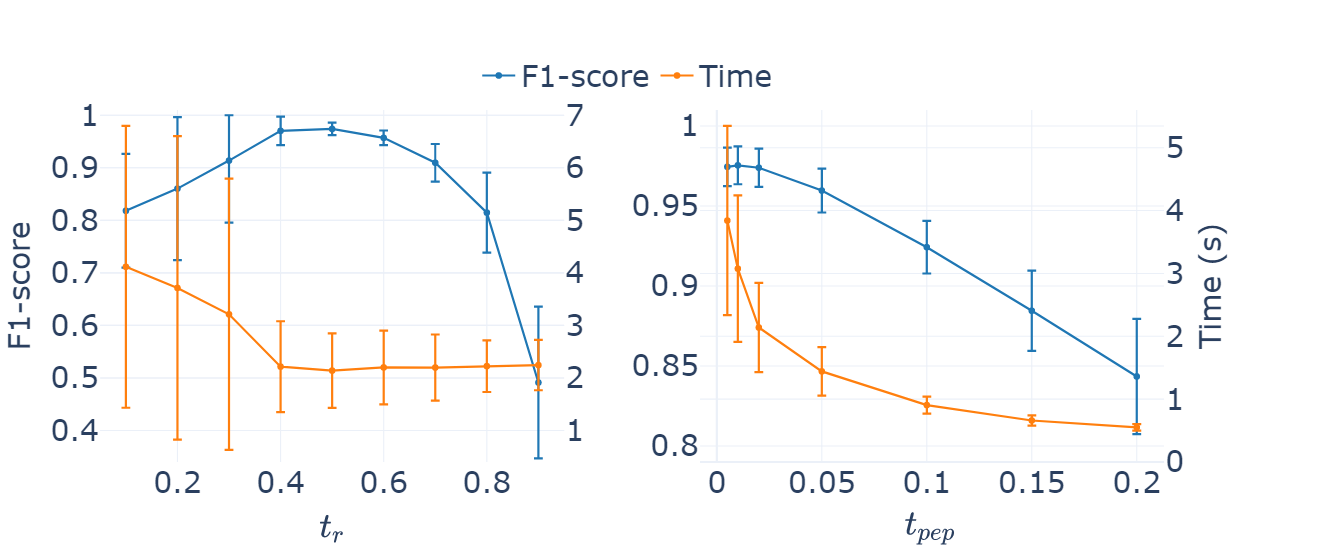}
    \captionof{figure}{Parameter sensitivity for risk threshold $t_r$ and proportional explanatory power threshold $t_{pep}$.}
    \label{fig:param_sens}
\end{figure}

We perform a parameter sensitivity analysis for RiskLoc. Two parameters can be tuned, the risk threshold $t_r$ and the proportional explanatory power threshold $t_{pep}$. We adjust them one by one while keeping the other fixed at the default value ($t_r=0.5$ and $t_{pep}=0.02$) using six datasets: $\mathcal{B}_i, i \in [0,..,4]$ and $\mathcal{D}$.
We report the average and standard deviation of the F1-scores and running times for each parameter setting and present the results in Figure~\ref{fig:param_sens}.

The performance is relatively stable for risk thresholds between $0.4$ and $0.6$. For lower risk thresholds, elements with a descendant element belonging to the root cause set may be returned erroneously. These parent elements incorporate the root cause, hence their $r_1$ score may be high but they simultaneously contain leaf elements that do not conform to the ripple effect. 
However, if $r_1$ is sufficiently high and $r_2$ does not compensate enough, then a low $t_r$ may return these elements. On the other hand, root cause elements may be missed if the risk threshold is too high.
We also note that the running time decreases with higher $t_r$ as fewer elements are considered and the search can be terminated earlier. 

For the proportional explanatory power threshold $t_{pep}$, we observe that both the F1-score and running time increase with lower values. With a lower threshold, the iterative search will terminate later, allowing for more calls to Algorithm~\ref{alg:element_search}, thus increasing the running time. Simultaneously, the performance will increase since fewer root cause elements are missed. However, a low $t_{pep}$ may return extraneous elements if there are elements within the abnormal partition that do not belong to the root cause set.
Performance stabilizes around $0.02$, while running time begins to increase exponentially for lower values.

\subsection{Ablation Study}
\label{sec:ablation}
We conduct an ablation study to validate the effectiveness of the main components of our proposed method. Specifically, we investigate the performance impact of removing or altering the following four integral components.

\begin{itemize}
    \item \textbf{No outlier removal}: No outliers removed in Algroithm~\ref{alg:weight_function}.
    \item \textbf{No $r_1$}: The $r_1$ component in the risk score is fixed to $1$.
    \item \textbf{No $r_2$}: The $r_2$ component in the risk score is fixed to $0$.
    \item \textbf{No weights}: The leaf element weights are removed (equivalent to setting all weights to $1$). 
\end{itemize}
The results are reported in Table~\ref{tab:ablation}. For each variant, the best values for the parameters $t_{pep}$ and $t_r$ are identified from the search space. 
We first make the wrap-up observation that all variants have lower or similar performance than RiskLoc on all datasets except for where no outliers are removed on the $\mathcal{L}$ dataset. This dataset has very few abnormal leaf elements; therefore, removing a small number of outliers can confuse the direction of the anomaly. A more complex outlier removal scheme could reduce the negative impact or could be avoided by assuming that the anomaly direction is given by the prior anomaly detection stage.

We further note that while each component is indispensable, the impact is highly dependent on the dataset characteristics. RiskLoc without outlier removal achieves markedly worse results on $\mathcal{A}$ and $\mathcal{A}_\star$ but has less import on the other datasets. On the other hand, $r_1$ is a key component on the $\mathcal{B}_i$ and $\mathcal{D}$ datasets, while $r_2$ and leaf element weights are essential on nearly all datasets. 

Furthermore, we study the effects of the element pruning scheme. The element pruning does not affect the F1-score but the running time. The results of no pruning are compared to pruning restricted to layer one and three are shown in Table~\ref{tab:ablation_time}. As can be observed, pruning will always decrease the running time. However, for smaller datasets, its advantageous to limit the pruning to the first layer, while for the largest dataset $\mathcal{H}$, allowing deeper element pruning is faster. This is due to the incurred overhead cost.

\begin{table}[t]
    \centering
    \small
    \caption{Ablation study. The relative worst F1-score is in bold for each variant.}
    \label{tab:ablation}
    \begin{tabular}{lccccccc}
         \toprule
         \multirow{2}*{\textbf{Variant}} & \multicolumn{7}{c}{\textbf{Dataset}} \\
         \cmidrule{2-8}
         & $\mathcal{A}$ & $\mathcal{B}_i$ & $\mathcal{D}$ & $\mathcal{A}_\star$ & $\mathcal{S}$ & $\mathcal{L}$ & $\mathcal{H}$ \\
         \toprule
         RiskLoc & $0.77$ & $0.98$ & $0.96$ & $0.46$ & $0.64$ & $0.68$ & $0.49$ \\
         \midrule
         No outlier removal & $\mathbf{0.22}$ & $0.98$ & $0.90$ & $0.28$ & $0.61$ & $0.75$ & $0.41$ \\
         No $r_1$ & $0.74$ &	$0.79$ &	$\mathbf{0.63}$ &	$0.44$ &	$0.63$ &	$0.48$ & $0.42$ \\ 
         No $r_2$ & $0.50$ &	$0.92$ &	$0.88$ &	$0.18$ &	$0.62$ &	$\mathbf{0.00}$ & $0.36$ \\
         No weights & $\mathbf{0.35}$ &$0.88$ &	$0.67$ &	$0.33$ &	$0.60$ &	$0.50$ & $0.33$ \\ 
         \bottomrule
    \end{tabular}
\end{table}

\begin{table}[t]
    \centering
    \small
    \caption{Average running time (s) with different element pruning layer restrictions.}
    \label{tab:ablation_time}
    \begin{tabular}{lccccccc}
         \toprule
         \multirow{2}*{\textbf{Variant}} & \multicolumn{7}{c}{\textbf{Dataset}} \\
         \cmidrule{2-8}
         & $\mathcal{A}$ & $\mathcal{B}_i$ & $\mathcal{D}$ & $\mathcal{A}_\star$ & $\mathcal{S}$ & $\mathcal{L}$ & $\mathcal{H}$ \\
         \toprule
         No pruning & $1.70$ & $2.27$ & $2.27$ & $4.05$ & $15.0$ & $3.37$ & $9{,}308$ \\
         \midrule
         Layer 1 & $1.20$ & $1.78$ & $1.76$ & $2.86$ & $12.0$ & $2.64$ & $6{,}838$ \\
         Reduction (\%) & $29.4$ & $21.6$ & $22.5$ & $29.4$ & $20.0$ & $21.7$ & $26.5$ \\
         \midrule
         Layer 3 & $1.52$ & $1.94$ & $2.08$ & $3.66$ & $12.5$ & $3.00$ & $5{,}959$ \\
         Reduction (\%) & $10.6$ & $14.5$ & $8.4$ & $9.6$ & $16.7$ & $11.0$ & $36.0$ \\
         \bottomrule
    \end{tabular}
\end{table}


\section{Conclusion}
In this paper, we propose RiskLoc, an efficient and accurate algorithm to locate root causes in multi-dimensional time series without significant root cause assumptions. Specifically, RiskLoc first divides the leaf elements into normal and abnormal partitions and assigns importance weights. All elements are evaluated using a risk score which integrates the weighted distribution of descending leaf elements and the impact of adjusting for the ripple effect. A set of root cause elements is identified by iteratively searching layer by layer until elements with a sufficiently large risk score and explanatory power are identified. 
We conduct extensive experiments on several semi-synthetic and three complementary synthetic datasets to verify the effectiveness and efficiency of our approach. We demonstrate that RiskLoc consistently outperforms state-of-the-art baselines, especially in challenging root cause scenarios.
In particular, on the most difficult semi-synthetic dataset, we achieve a relative increase in F1-score of 57\% compared to the second-best approach, while gains of up to 41\% are obtained on the synthetic datasets.
For future work, we will focus on enhancing the model robustness in scenarios with low quality or missing forecast values, and further explore new partitioning schemes and innovative pruning strategies.

\bibliographystyle{ACM-Reference-Format}
\bibliography{references}

\clearpage

\appendix
\section{Synthetic Dataset Details}
\label{app:synthetic}
As introduced in Section~\ref{sec:experimental_setup}, we generate three synthetic datasets $\mathcal{S}$, $\mathcal{L}$, and $\mathcal{H}$ to use in our method evaluation. Dataset $\mathcal{S}$ has $1{,}000$ instances, each with $5$ attributes with value sets of size $10$, $12$, $10$, $8$, and $5$, respectively, for a total of $48{,}000$ leaf elements. For dataset $\mathcal{L}$, we generate $1000$ instance with $4$ dimensions of size $10$, $24$, $10$, and $15$, i.e., a total of $36{,}000$ leaf elements. Finally, $\mathcal{H}$ contains $100$ instances, each with $6$ dimensions of size $10$, $5$, $250$, $20$, $8$, and $12$, for a total of $24{,}000{,}000$ leaf elements.

For the forecast residuals in Equation~\ref{eq:synth_forecast}, we use $\sigma \sim U[0, 0.25]$ for $\mathcal{S}$ and $\mathcal{H}$, and $\sigma \sim U[0, 0.10]$ for $\mathcal{L}$. The actual and forecast values are then randomly swapped for each element with a 50\% probability to ensure that the noise is equally distributed.  

We introduce synthetic anomalies to the data where the number of anomalies for each instance is randomly drawn from a range. 
For $\mathcal{S}$ and $\mathcal{H}$, we use $[1,3]$ while $\mathcal{L}$ use $[1,5]$, i.e., $\mathcal{L}$ can have up to $5$ anomalies in a single instance.
The number of elements in each anomaly is likewise uniformly random between $[1,3]$ for $\mathcal{S}$ and $\mathcal{H}$, while $\mathcal{L}$ is fixed to $1$ since the anomalies in $\mathcal{L}$ are only placed in the highest layer.
We make sure that there are no overlapping anomalies, e.g., if one anomaly is in element $(*,*,c5,*)$ then an anomaly $(*,*,c5,d3)$ is not allowed. We also ensure that each cuboid contains only a single anomaly (although it can involve multiple elements).

Each anomaly has a severity $s$ and a deviation $d$ corresponding to its distance from the normal samples and the variance of its leaf elements. We set $s \sim U[0.25, 1.0]$ and $d \sim U[0.0, 0.1]$ for $\mathcal{S}$ and $\mathcal{H}$, and $s \sim U[0.5, 1.0]$ and $d=0.0$ for $\mathcal{L}$. $\mathcal{L}$ does not require any deviation since each anomaly is a single leaf element. We allow for the anomaly direction to change (i.e., if $\sum{f(e)} > \sum{v(e)}$ or vice versa) but for a single generated instance, the direction of the anomalies are the same. All elements in a single anomaly are scaled the same (i.e., following the \textit{ripple effect}~\cite{hotspot2018}) with:
\begin{equation}
    x = \max(x \cdot (1-N(s, d)), 0)
\end{equation}
where $x=v(e)$ if $\sum{v(e)} > \sum{f(e)}$, otherwise $x=f(e)$. 

\section{Experimental Details}
\label{app:exp_details}
We run all experiments on a Linux server with 24 x Intel(R) Xeon(R) CPU E5-2620 v3 @ 2.40GHz and 128G RAM memory. All algorithms are implemented in Python and rely heavily upon the Pandas and NumPy libraries. The experiments were all carried out under the same conditions and parallelization is done over the instances (files) themselves. Notably, by adjusting the code to run the cuboids (or clusters for Squeeze and AutoRoot) in parallel, a significant reduction in running time could be obtained.

For each baseline algorithm, we try a set of values for each tunable parameter and select the ones that achieve the highest average F1-score on the $\mathcal{B}_i, i \in [0,..,4]$ datasets. We define $S_1=\{ i/10, i \in [1,..,9] \}$ and $S_2=S_1 \cup \{0.01, 0.05, 0.15\}$.  

\noindent
\textbf{Adtributor}: We run the algorithm with $T_{EP} \in S_1$ and $T_{EEP} \in S_2$ with the condition $T_{EP} \geq T_{EEP}$. The optimal parameters were determined to be $T_{EP}=0.1$ and $T_{EEP}=0.1$ which are used in all experiments. $k$ is set to $3$ for all datasets since it is the highest number of anomalies that can be found. We tried with $k=5$ for $\mathcal{A}_\star$ and $\mathcal{S}$ but did not get better results.\\ 

\noindent
\textbf{R-Adtributor}: Similar to the standard Adtributor algorithm, we use $T_{EEP} \in S_2$ and select the best $T_{EEP}$ value. The value of $T_{EEP}$ is set to $0.2$ and $k$ is set to $3$. \\

\noindent
\textbf{HotSpot}:
For HotSpot, we try using both the original score, $ps$~\cite{hotspot2018}, and the extended $GPS$~\cite{squeeze2019} with the threshold $PT \in S_1$. The best result is achieved by using $GPS$ with $PT=0.8$. We further set $M=200$ which is a trade-off between running time and performance. In the original article, $M$ is set to values between $5$ and $15$. Here, we allow for larger values, which may give higher F1-scores at the cost of longer running times. A higher value for $M$ may give better results, however, HotSpot with $M=200$ is already requires more than twice the running time as compared to the second slowest algorithm on most datasets. \\ 

\noindent
\textbf{Squeeze}:
The code for Squeeze has been open sourced.\footnote{\url{https://github.com/NetManAIOps/Squeeze}} We use their implementation with default values for all experiments with the exception of setting the maximum returned element per cluster for $\mathcal{H}$ to 3 (the optimal value) to improve the running speed. Note that the code was revised following the "known issues" as indicated in the README.md file to enhance the performance and uses the amended formula for constant C\footnote{\url{https://github.com/NetManAIOps/Squeeze/issues/6}} as follows: 
\begin{align}
\begin{split}
    g_{cluster}   &= \frac{\log(num\_cluster + 1)}{num\_cluster} \\
    g_{attribute} &= \frac{num\_attr}{\log(num_attr + 1)} \\
    g_{coverage}  &= -\log(\text{coverage of abnormal leaves}) \\
    C             &= g_{cluster} \cdot g_{attribute} \cdot g_{coverage}
\end{split}
\end{align} \\

\noindent
\textbf{AutoRoot}:
AutoRoot has a single parameter $\delta$. We tried a set of values $\{ 0.5, 0.3, 0.25, 0.2, 0.1, 0.05, 0.01 \}$ and found that $0.25$ performs the best on the $\mathcal{B}_i$ datasets. However, we noticed that the performance on the more challenging datasets is markedly worse than for a lower $\delta$. To get a fairer baseline we therefore set $\delta=0.1$ on $\mathcal{A}$, $\mathcal{A}_\star$ and the synthetic datasets.\\

\noindent
\textbf{RiskLoc}:
We run RiskLoc with $t_r=0.5$ and $t_{pep}=0.02$. The details of the parameter sensitivity are in Section~\ref{sec:param_sens}.
The reported results for $\mathcal{D}$ is using the original calculation for explanatory power and not the extended formula for derived measures~\cite{adtributor2014}, increasing the F1-score by around 7\% on average. In the running-time experiments, we have restricted element pruning to the first layer for all datasets except $\mathcal{H}$ which use up to the third layer.


\section{Additional Experimental Results}
\label{app:add_exp_res}
We present the full results of the algorithms on all root cause scenarios for the datasets $\mathcal{A}$, $\mathcal{B}_i$ with $i \in [0,..,4]$, and $\mathcal{D}$ in Table~\ref{tab:full_result} and dataset $\mathcal{A}_\star$ in Table~\ref{tab:results_A_star}. In both tables, layer denotes the layer with the ground-truth root causes while \#elements signify the number of elements within the root causes.

\begin{table*}[t]
    \centering
    \small
    \caption{F1-score comparison.}
    \label{tab:full_result}
    \begin{tabular}{clccccccccc}
         \toprule
         
         \multirow{2}*{\textbf{Dataset}} &  \multirow{2}*{\textbf{Algorithm}} & \multicolumn{9}{c}{\textbf{Root cause (layer, \#elements)}} \\
         \cline{3-10}
         & & (1,1) & (1,2) & (1,3) & (2,1) & (2,2) & (2,3) & (3,1) & (3,2) & (3,3) \\
         \toprule
         \multirow{7}{*}{$\mathcal{A}$}  & Adtributor & 0.5005 & 0.4059 & 0.3445 & 0.0000 & 0.0000 & 0.0000 & 0.0000 & 0.0000 & 0.0000 \\
          & R-Adtributor & 0.0404 & 0.0052 & 0.0015 & 0.0086 & 0.0042 & 0.0010 & 0.0000 & 0.0000 & 0.0000 \\
          & HotSpot & 0.2765 & 0.1537 & 0.0920 & 0.0415 & 0.0251 & 0.0146 & 0.0016 & 0.0004 & 0.0000 \\
          & Squeeze & 0.8882 & 0.7719 & 0.6476 & 0.8102 & 0.6470 & 0.5275 & 0.4992 & 0.4089 & 0.3592 \\
          & AutoRoot & 0.2654 & 0.1728 & 0.1393 & 0.4131 & 0.2451 & 0.2035 & 0.5136 & 0.3610 & 0.2452 \\
          & RiskLoc & \textbf{0.9997} & \textbf{0.8097} & \textbf{0.6838} & \textbf{0.9023} & \textbf{0.7771} & \textbf{0.6701} & \textbf{0.7948} & \textbf{0.6714} & \textbf{0.6083} \\
          
         \midrule
         \multirow{7}{*}{$\mathcal{B}_0$} & Adtributor & 0.6494 & 0.5522 & 0.4682 & 0.0000 & 0.0000 & 0.0000 & 0.0000 & 0.0000 & 0.0000 \\
         & R-Adtributor & 0.4930 & 0.2213 & 0.0647 & 0.1739 & 0.1382 & 0.0824 & 0.0401 & 0.0684 & 0.0595 \\
         & HotSpot & 0.9600 & 0.6267 & 0.5025 & 0.8600 & 0.5533 & 0.4561 & 0.5100 & 0.4667 & 0.4000 \\
         & Squeeze & 0.8811 & 0.9448 & 0.9640 & 0.9565 & \textbf{0.9875} & 0.9352 & 0.9327 & 0.9474 & \textbf{0.9538} \\
         & AutoRoot & \textbf{1.0000} & 0.9772 & 0.9130 & \textbf{0.9900} & 0.9561 & 0.8588 & \textbf{0.9900} & 0.9344 & 0.8079 \\
         & RiskLoc & 0.9600 & \textbf{0.9700} & \textbf{0.9775} & 0.9286 & 0.9723 & \textbf{0.9592} & 0.9744 & \textbf{0.9556} & 0.9437 \\
         
         \midrule
         \multirow{7}{*}{$\mathcal{B}_1$} & Adtributor & 0.6557 & 0.6212 & 0.4887 & 0.0000 & 0.0000 & 0.0000 & 0.0000 & 0.0000 & 0.0000 \\
         & R-Adtributor & 0.6019 & 0.2458 & 0.0712 & 0.1778 & 0.1538 & 0.0712 & 0.0346 & 0.0288 & 0.0511 \\
         & HotSpot & 0.9600 & 0.6200 & 0.4840 & 0.8800 & 0.5867 & 0.4100 & 0.4000 & 0.4000 & 0.3300 \\
         & Squeeze & 0.9333 & 0.9610 & 0.9866 & 0.8732 & 0.9455 & 0.9231 & 0.8815 & 0.9389 & 0.9325 \\
         & AutoRoot & 0.9852 & \textbf{0.9875} & 0.9364 & 0.9901 & 0.9745 & 0.9117 & 0.9515 & 0.9330 & 0.7992 \\
         &  RiskLoc & \textbf{0.9950} & 0.9850 & \textbf{0.9898} & \textbf{1.0000} & \textbf{0.9925} & \textbf{0.9651} & \textbf{0.9899} & \textbf{0.9590} & \textbf{0.9640} \\
         
         \midrule
         \multirow{7}{*}{$\mathcal{B}_2$} & Adtributor & 0.6689 & 0.6340 & 0.4615 & 0.0000 & 0.0000 & 0.0000 & 0.0000 & 0.0000 & 0.0000 \\
         & R-Adtributor & 0.6207 & 0.2318 & 0.1100 & 0.1654 & 0.1511 & 0.1081 & 0.0118 & 0.0418 & 0.0553 \\
         & HotSpot & 0.9900 & 0.6600 & 0.4350 & 0.8800 & 0.6107 & 0.4110 & 0.4000 & 0.3467 & 0.3300 \\
         & Squeeze & 0.8558 & 0.9746 & 0.9420 & 0.8889 & 0.8244 & 0.8174 & 0.8835 & 0.8655 & 0.8571 \\
         & AutoRoot & \textbf{1.0000} & 0.9872 & 0.9664 & \textbf{0.9899} & 0.8732 & 0.8024 & 0.9255 & 0.8433 & 0.7230 \\
         & RiskLoc & \textbf{1.0000} & \textbf{0.9899} & \textbf{0.9774} & 0.9612 & \textbf{0.9924} & \textbf{0.9882} & \textbf{0.9848} & \textbf{0.9848} & \textbf{0.9726} \\
         
         \midrule
         \multirow{7}{*}{$\mathcal{B}_3$} & Adtributor & 0.6579 & 0.6611 & 0.4988 & 0.0000 & 0.0000 & 0.0000 & 0.0000 & 0.0000 & 0.0000 \\
         & R-Adtributor & 0.6634 & 0.3361 & 0.1852 & 0.1139 & 0.1871 & 0.1101 & 0.0227 & 0.0479 & 0.0337 \\
         & HotSpot & 1.0000 & 0.6622 & 0.4650 & 0.9000 & 0.5800 & 0.4300 & 0.3200 & 0.3267 & 0.2957 \\
         & Squeeze & 0.8952 & 0.9347 & 0.8897 & 0.8349 & 0.8173 & 0.7692 & 0.7798 & 0.7910 & 0.7701 \\
         & AutoRoot & \textbf{1.0000} & 0.9819 & 0.9131 & 0.9381 & 0.8981 & 0.7854 & 0.9101 & 0.8840 & 0.8311 \\
         & RiskLoc & \textbf{1.0000} & \textbf{0.9924} & \textbf{0.9686} & \textbf{1.0000} & \textbf{0.9975} & \textbf{0.9883} & \textbf{0.9848} & \textbf{0.9692} & \textbf{0.9565} \\

         \midrule
         \multirow{7}{*}{$\mathcal{B}_4$} & Adtributor & 0.6623 & 0.6954 & 0.5263 & 0.0000 & 0.0000 & 0.0000 & 0.0000 & 0.0000 & 0.0000 \\
         & R-Adtributor & 0.7488 & 0.3730 & 0.1143 & 0.0735 & 0.1135 & 0.1073 & 0.0220 & 0.0377 & 0.0450 \\
         & HotSpot & 0.9900 & 0.6312 & 0.3800 & 0.9100 & 0.6020 & 0.4450 & 0.2400 & 0.2533 & 0.2807 \\
         & Squeeze & 0.8531 & 0.8732 & 0.8430 & 0.8224 & 0.7196 & 0.7097 & 0.7207 & 0.7538 & 0.7206 \\
         & AutoRoot & 0.9950 & 0.9561 & 0.8963 & 0.9082 & 0.8490 & 0.7582 & 0.7708 & 0.7608 & 0.7107 \\
         & RiskLoc & \textbf{1.0000} & \textbf{0.9826} & \textbf{0.9607} & \textbf{1.0000} & \textbf{0.9975} & \textbf{0.9698} & \textbf{0.9744} & \textbf{0.9558} & \textbf{0.9383} \\
         \midrule
         \multirow{6}{*}{$\mathcal{D}$} & Adtributor & 0.5556 & 0.5835 & 0.4979 & 0.0000 & 0.0000 & 0.0000 & 0.0000 & 0.0000 & 0.0000 \\
         & R-Adtributor & 0.3247 & 0.2824 & 0.1262 & 0.0194 & 0.0672 & 0.1114 & 0.0021 & 0.0080 & 0.0079 \\
         & HotSpot & 0.9000 & 0.6667 & 0.5215 & 0.8800 & 0.6533 & 0.5013 & 0.9000 & 0.6667 & 0.5000 \\
         & Squeeze & \textbf{0.9500} & \textbf{0.9850} & \textbf{0.9760} & 0.9500 & 0.9500 & \textbf{0.9449} & 0.9254 & 0.8750 & 0.8848 \\
         & AutoRoot & 0.6929 & 0.7906 & 0.7871 & 0.6071 & 0.7621 & 0.7721 & 0.6873 & 0.7723 & 0.7129 \\
         & RiskLoc & \textbf{0.9500} & 0.9825 & 0.9666 & \textbf{0.9548} & \textbf{0.9698} & 0.9454 & \textbf{0.9543} & \textbf{0.9622} & \textbf{0.9370} \\
         
         \bottomrule
    \end{tabular}
\end{table*}

\begin{table*}[t]
    \centering
    \small
    \caption{F1-score comparison on $\mathcal{A}_\star$.}
    \label{tab:results_A_star}
    \begin{tabular}{lccccccccc}
         \toprule
         \multirow{2}*{\textbf{Algorithm}} & \multicolumn{9}{c}{\textbf{Root cause (layer, \#elements)}} \\
         \cline{2-10}
         & (1,4) & (1,5) & (2,4) & (2,5) & (3,4) & (3,5) & (4,1) & (4,2) & (4,3) \\
         \toprule
         Adtributor & 0.3062 & 0.2790 & 0.0000 & 0.0000 & 0.0000 & 0.0000 & 0.0000 & 0.0000 & 0.0000 \\
         R-Adtributor & 0.0000 & 0.0000 & 0.0003 & 0.0000 & 0.0000 & 0.0000 & 0.0000 & 0.0000 & 0.0000 \\
         HotSpot & 0.0618 & 0.0384 & 0.0097 & 0.0033 & 0.0008 & 0.0000 & 0.0000 & 0.0000 & 0.0000 \\
         Squeeze & 0.5463 & 0.4686 & 0.4219 & 0.3566 & 0.2213 & 0.2158 & 0.2286 & 0.1278 & 0.0693 \\
         AutoRoot & 0.1284 & 0.1753 & 0.1820 & 0.2379 & 0.2515 & 0.2549 & 0.0699 & 0.1636 & 0.1586 \\
         RiskLoc & \textbf{0.5644} & \textbf{0.4788} & \textbf{0.5710} & \textbf{0.4487} & \textbf{0.5179} & \textbf{0.4754} & \textbf{0.5188} & \textbf{0.3472} & \textbf{0.2632} \\
         \bottomrule
    \end{tabular}
\end{table*}

\end{document}